\documentclass[a4paper,conference]{IEEEtran}

\usepackage{subfigure}

\usepackage{graphicx}
\usepackage{comment}
\usepackage{amsmath,amssymb} 

\usepackage{color}
\usepackage{soul} 

\usepackage{balance} 

\usepackage[normalem]{ulem} 

\ifCLASSINFOpdf
\else
\fi
\begin{document}
%
\title{Context Matters: Self-Attention for\\Sign Language Recognition}

\author{\IEEEauthorblockN{Fares Ben Slimane}
\IEEEauthorblockA{Department of Computer Science\\ University of Quebec at Montreal\\
Montreal, Quebec, Canada\\
ben\_slimane.fares@courrier.uqam.ca}
\and
\IEEEauthorblockN{Mohamed Bouguessa}
\IEEEauthorblockA{Department of Computer Science\\ University of Quebec at Montreal\\
Montreal, Quebec, Canada\\
bouguessa.mohamed@uqam.ca}
}


%


\maketitle

\begin{abstract}

This paper proposes an attentional network for the task of Continuous Sign Language Recognition. The proposed approach exploits co-independent streams of data to model the sign language modalities. These different channels of information can share a complex temporal structure between each other. For that reason, we apply attention to synchronize and help capture entangled dependencies between the different sign language components. Even though Sign Language is multi-channel, handshapes represent the central entities in sign interpretation. Seeing handshapes in their correct context defines the meaning of a sign. Taking that into account, we utilize the attention mechanism to efficiently aggregate the hand features with their appropriate spatio-temporal context for better sign recognition. We found that by doing so the model is able to identify the essential Sign Language components that revolve around the dominant hand and the face areas. We test our model on the benchmark dataset RWTH-PHOENIX-Weather 2014, yielding competitive results. 


\end{abstract}


%
\IEEEpeerreviewmaketitle

\section{Introduction}

Sign languages are often defined as manual languages. However, besides the hand articulations, non-manual components like facial expressions, arm, head, body movements and positions play a crucial part in Sign Languages \cite{hand1}. Any change in one of these components can alter the meaning of a sign. Usually, the handshape performed by the dominant hand carries most of the meaning of the sign \cite{hand2}. As for the other components, they provide a semantic context to it. For example, the signs for \textit{``man"} and \textit{``woman"} are very close as they both use the same handshape and body movement, although, as mentioned in \cite{omniglot}, male words tend to be signed close to the forehead while female signs are usually performed around the chin. Similarly, for the signs \textit{``sit"} and \textit{``chair"}, they both have the same handshape but slightly different hand movements, in which \textit{``chair"} is the sign \textit{``sit"} but twice \cite{valli2000linguistics}. Facial expressions can also greatly affect the meaning of a handshape, and are mostly used to ask questions or express emotion. For instance, when signing a \textit{``you"} sign, by raising the eyebrows, it becomes a question: \textit{``Is it you?"} or by using a different facial expression it turns into \textit{``It's you?!?!"} meaning that \textit{``I am surprised that it is you"} \cite{lifeprint}. These examples highlight the idea that interpreting a sign often requires recognizing the handshape accompanied by its contextual information. This context is not limited to the spatial information that resides around the handshape for a fixed point in time, but also the temporal information that consists of hand and body movements. 

Much of the current literature in Continuous Sign Language Recognition (CSLR) ignores the notion of incorporating contextual information to the handshape. Instead, they either exclusively use the handshape features from the cropped hand images \cite{koller2016deep}, or simply use global features by trying to learn a complete spatial representation of full-body images~\cite{koller2017re,wu2016deep}. Nevertheless, there have been some works that investigated the effect of combining information of the handshape with the other Sign Language modalities. For instance, \cite{huang2018video} use a two-stream 3D-CNN to extract global (full-body) and local (hand) features and then fuse the representations in the last fully connected layers. Another work \cite{camgoz2017subunets} involves the use of specialized sub-networks (SubUNets) to learn full-frame and handshape information and then to synchronize the modalities through a recurrent BLSTM layer. The authors found that combining modalities in this manner outperforms the use of isolated SubUNets.
However, it is important to note that sign language modalities mostly share complex non-linear relations. Because of that, these approaches may fail to successfully capture and aggregate the required handshape dependencies. This is especially true when using recurrent networks, as they are more biased by close contexts, only considering short-term dependencies rather than the global long-term dependencies \cite{bengio1994learning}. In this paper, we propose a superior method that efficiently combines handshape features with their appropriate global context, and which yields better recognition results than the latter studies. 

Virtually all progress in CSLR has operated under the assumption that this is a gesture recognition task and that the main challenge is to simply learn the mapping of signs to their respective glosses. The authors in \cite{cihan2018neural} argue that this is not the case, as Sign Language has its own linguistic and grammatical properties, and that we should take into account the word orders and grammar. That being the case, they introduce a new problem setup: Sign Language Translation (SLT)\footnote{We draw the attention of the reader that in this study we tackle the specific problem of Continuous Sign Language Recognition (CSLR). The task of Sign Language Translation (SLT) is beyond the scope of this paper.}, in which they approach it as a machine translation task, generating spoken language translations from sign language videos. And they, accordingly, employ an encoder-decoder RNN with attention which is commonly used for machine translation tasks. Alternatively, it has been shown that using an attention network by itself can also be efficient in machine translation. The Transformer Network from \cite{vaswani2017attention}, which is based solely on attention, has been proven to be very effective in machine translation tasks. Just recently, Camgoz et al. \cite{camgoz2020} proposed a transformer-based encoder-decoder architecture for the task of SLT which significantly exceeds the previous state-of-the-art performance from \cite{cihan2018neural} and sets a new baseline result for SLT. This proves the efficiency of such architecture for capturing temporal dependencies. 

Unlike spoken languages, sign language is undoubtedly multi-channel. Information is carried through handshapes, body movements, and facial expressions. Contrary to \cite{camgoz2020} and instead of relying on a unique channel of information from full-frame representations, we exploit the attention mechanism to effectively combine different modality information. When signing, the handshape usually depends on some relevant contexts, rather than all context information. As a result, the transformer network is a more suitable choice for our problem setup, as it can efficiently combine hand features with their appropriate full-body information, since it is explicitly built to detect important dependencies, as opposed to their recurrent counterpart.

Accordingly, in this paper, we propose an attention-based approach for sequence to sequence sign language alignment and recognition. Unlike previous works, the originality of our approach lies in explicitly picking up and aggregating contextual information from the non-manual sign language components. Without any domain annotation, our approach is able to exclusively identify the most relevant features associated with the handshape when predicting a sign. The main contributions of this paper can be summarized as follows:   

\begin{itemize}

\item{Devising an end-to-end framework for sequence to sequence Sign Language Recognition that utilizes self-attention for temporal modeling. }

\item{Elaborating a more efficient method to incorporate handshapes with their spatiotemporal context for Sign Language Recognition.}

\item{Achieving competitive results, in terms of Word Error Rate, on the RWTH-PHOENIX-Weather 2014 benchmark dataset.}

\end{itemize}

\section{Related work}

Most work in CSLR can be mainly divided into two stages: the extraction of spatial representations from the input images and then learning the correspondence between the representations' sequence and the target sequence. There exists a fair amount of literature in the field that utilizes handcrafted feature representations \cite{monnier2014multi} and classical models like HMMs for sequence learning \cite{koller2016deep}. However, due to the revolutionary advances in deep learning, recent research has started to shift interest to using CNNs \cite{koller2016deep}, and 3D-CNNs \cite{huang2018video} as feature extractors, and recurrent networks for temporal modelling \cite{camgoz2017subunets}. For instance, the authors in \cite{koller2017re} propose a CNN-RNN-HMM architecture for the task of CSLR. On top of their CNN-RNN-HMM model, \cite{koller2017re} employ an iterative forced alignment algorithm that refines the label-to-image alignment and that broadly outperforms the previous work by a large margin. Instead of manual labelling, forced alignment is able to provide useful training targets, which are re-fined through iterative alignment. The model will be trained on these new labels and re-align them for the next iteration. The training process is repeated until the resulting model stops getting better recognition. A similar work \cite{pami2019} employ the same forced alignment technique to maximize the target likelihood estimation. The authors adopt a multi-stream HMM approach to synchronize and learn from parallel and different sub-problems (sign language, mouth shape, and hand shape recognition). Respectively, using a parallel alignment approach produces better recognition performance than \cite{koller2017re} which follows a similar CNN-LSTM-HMM setup but uses a single stream input.

As a rule, CSLR is considered a weakly supervised problem since gloss annotations usually lack the exact temporal location \cite{Cui_2017_CVPR}. Besides HMMs, a common strategy used to solve this problem is the use of Connectionist Temporal Classification (CTC) loss function \cite{graves2006connectionist}, since it considers all possible alignments between input and output sequences. CTC has been successfully applied for CSLR \cite{camgoz2017subunets,Cui_2017_CVPR} and also for various other sequence to sequence problems including speech recognition \cite{graves2013speech} and handwriting recognition \cite{graves2008novel}.
Nonetheless, the CTC algorithm is found to be incompatible for machine translation since it only allows monotonic alignment between source and target sequences \cite{hannun2017sequence}. Instead, the encoder-decoder architecture is usually applied for machine translation. As described in \cite{cho2014learning}, the encoder-decoder network consists of learning a latent representation that maps the input sequence to another target sequence. However, this usually results in having a bottleneck state that fails to capture long-range dependencies, especially when dealing with long input sequences. Giving a network an attention mechanism is widely considered to be a good way to solve this problem, as it allows the decoder to focus on relevant hidden states, and avoids squeezing all input information into the last hidden state \cite{bahdanau2014neural}. The authors in \cite{cihan2018neural} adopt this strategy for Sign Language Translation. They found that having an attention mechanism \cite{bahdanau2014neural} exceptionally improved the translation performance.
	
Recent machine translation research \cite{vaswani2017attention} achieved state of the art performance by merely using self-attention, entirely replacing recurrent models.
This attention mechanism has been successfully used in a variety of other tasks like video captioning \cite{chen2018tvt} in which the authors use an encoder-decoder framework with the transformer network backbone to generate a text description of a given video. Another interesting work in the field of reading comprehension \cite{yu2018qanet}, utilizes the self-attention mechanism to compute the similarities between a pair of context and query words. 

A separate study in action recognition \cite{girdhar2019video} proposes a transformer-based model. They argue that human actions are recognizable from the state of the environment, apart from their own pose. For that, they use self-attention to aggregate features from the spatio-temporal context around the person to correctly classify a person's actions. This inspires us to exploit, in this paper, self-attention to incorporate handshape features with their spatio-temporal dependencies. Doing so would add context from the global information to the handshape and ultimately contribute to improving sign recognition. To the best of our knowledge, no previous study has considered doing this for Sign Language Recognition.

The field of Continuous Sign Language Recognition only really took off in recent years, as labeled datasets became more publicly accessible for researchers. For instance, RWTH-PHOENIX-Weather 2014 \cite{koller2015continuous} that quickly became a popular baseline for CSLR. The dataset offers sign language gloss annotation for a German weather forecast. In this paper, we demonstrate the effectiveness of our approach through the use of the latter dataset.	

\section{Proposed Approach}

In this section, we describe the approaches that we have devised for the task of Continuous Sign Language Recognition (CSLR). First, in Section 3.B, we introduce our proposed approach which we refer to as Sign Attention Network (SAN). The considered model is based on the Transformer architecture that we briefly describe in Section 3.A. Our model first extracts spatial features from the sign clip frames using 2D CNNs. Then, it employs self-attention for temporal modeling. Finally, the proposed model learns sequence alignment through a CTC layer. In Section 3.C, we add a secondary stream for the cropped handshape sequences and we combine the hand features with their spatio-temporal full-body context. In Section 3.D, instead of considering the entire context information, we merely attend to information from the handshape local surroundings. This will allow the model to only focus on the required context, discarding unnecessary distant information. We explain the motivation behind such an approach and demonstrate its effectiveness in the experimentation section through quantitative and qualitative findings. Implementation and Training details are given in Sections 3.E and 3.F respectively.

\subsection{Transformer Network}
	
This architecture was firstly proposed in \cite{vaswani2017attention} as an alternative for the traditional recurrent models. It relies on self-attention to compute sequence representation, entirely repealing recurrence and convolutional operations, in which it helped reduce computational cost. Instead of performing attention on all the feature space, features are first linearly projected to query $Q$, key $K$, and value $V$ embeddings with lower dimensionality, $h$ times. According to the authors \cite{vaswani2017attention}, this allows the model to attend to information from different representation subspaces. The architecture then uses the scaled dot product (see Equation \ref{eq:1}) to calculate the attention scores (similarity) of each feature with the rest of the features in the sequence by multiplying $Q$ with $K$. The output is computed as the weighted sum of values $V$ conditionally to the attention weights. 
	
\begin{equation}
Attention(Q,K,V) = softmax\Big(\frac{QK^{T}}{\sqrt{d_k}}\Big)V
\label{eq:1}
\end{equation}
	
Moving away from recurrent and convolutional models, we end up losing positional information. Accordingly, to preserve the sequence order, the authors add positional encodings to the feature representations through the use of a sinusoid-wave-based function.
A detailed explanation of the overall architecture can be found in the original paper \cite{vaswani2017attention}.

\begin{figure*}[t]
\centering
\includegraphics[width=5in]{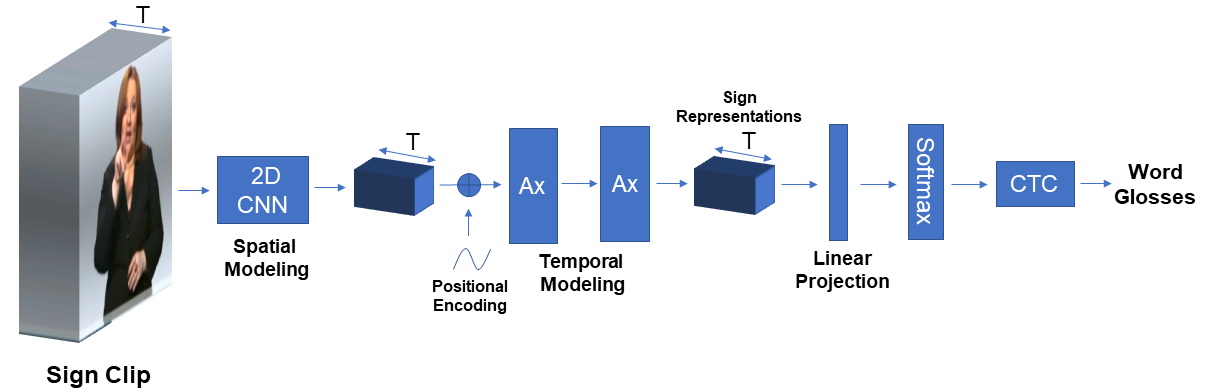}
\caption{Overview of our Sign Attention Network that takes a sequence of full-frame images and outputs the target word glosses. The Ax Unit represents the Attention stack as introduced in \cite{vaswani2017attention}, which is composed of a multi-head self-attention mechanism followed by a fully connected layer. We apply a layer Norm \cite{ba2016layer} and then a residual connection for each as opposed to the original Transformer paper.}
\label{fig:1}
\end{figure*}

\subsection{Sequence Alignment with CTC}
    
We propose an attention-based model for Sign Recognition. Our model is designed to accept a clip as input and accordingly produce the output words in a spoken language, making it a sequence to sequence task. Subsequently, our model is trained to recognize and time-align the target glosses from the sign language clip.

In recognition tasks, framewise aligning each timestep in the input sequence $X = \{x_1,x_2,..x_T\}$ with its corresponding ground-truth annotation can be time-consuming and computationally expensive to train. This is especially true in the case of CSLR as the input sequence's length can be much more superior than that of the target sequence. Connectionist Temporal Classification (CTC) proposed by \cite{graves2006connectionist}, is a popular solution to overcome such a problem. The CTC considers all possible alignment in each timestep. It introduces an extended target vocabulary $L'= L \cup \{\epsilon\}$ where $\epsilon$ represents the blank label that accounts for silence and which is to be removed from the output while decoding. 

The CTC objective loss function is defined as $\mathcal{L}_{CTC} = -\log  P(Y|X)$ where $P(Y|X)$ is the conditional probability and can be described as:

\begin{equation}
P(Y|X) = \sum_{a\in L'} \prod_{_{1\leq t\leq T}}
P_t(a_t|X)
\label{eq:2}
\end{equation}

Computing the score for each alignment can be computationally intensive; to that end, CTC uses a dynamic programming algorithm to efficiently and promptly compute $\mathcal{L}_{CTC}$ \cite{hannun2017sequence}. 

Recurrent networks are usually used to estimate posteriors $P_t(a_t|X)$ for each timestep $t_{1\leq t\leq T}$. However, RNNs normally fail to capture global dependencies, especially when dealing with long sequences. Therefore, for sequence to sequence modeling, we replace traditional recurrent methods with self-attention, as they have proven to be more resilient to the vanishing gradient problem. Similarly, as used in the encoder side of the Transformer network, we use stacked self-attention layers to learn temporal dependencies and map the input frame features viewed as  $X \in \mathbb{R}^{T\mathsf{x}D}$ 
to another sequence of equal length, in which $D$ represents the feature dimensionality. In summary, our model produces image representations through a 2D CNN on the clip individual frames of a given sign clip. The resulting matrix is passed to succeeding Ax Attention units followed by a linear projection and a Softmax to output the word probabilities and finally a CTC layer in order to generate the gloss words, as highlighted in Figure \ref{fig:1}.

\subsection{Context-Hand Attention Layer}

\begin{figure*}[t]
 \centering
 \includegraphics[width=6in]{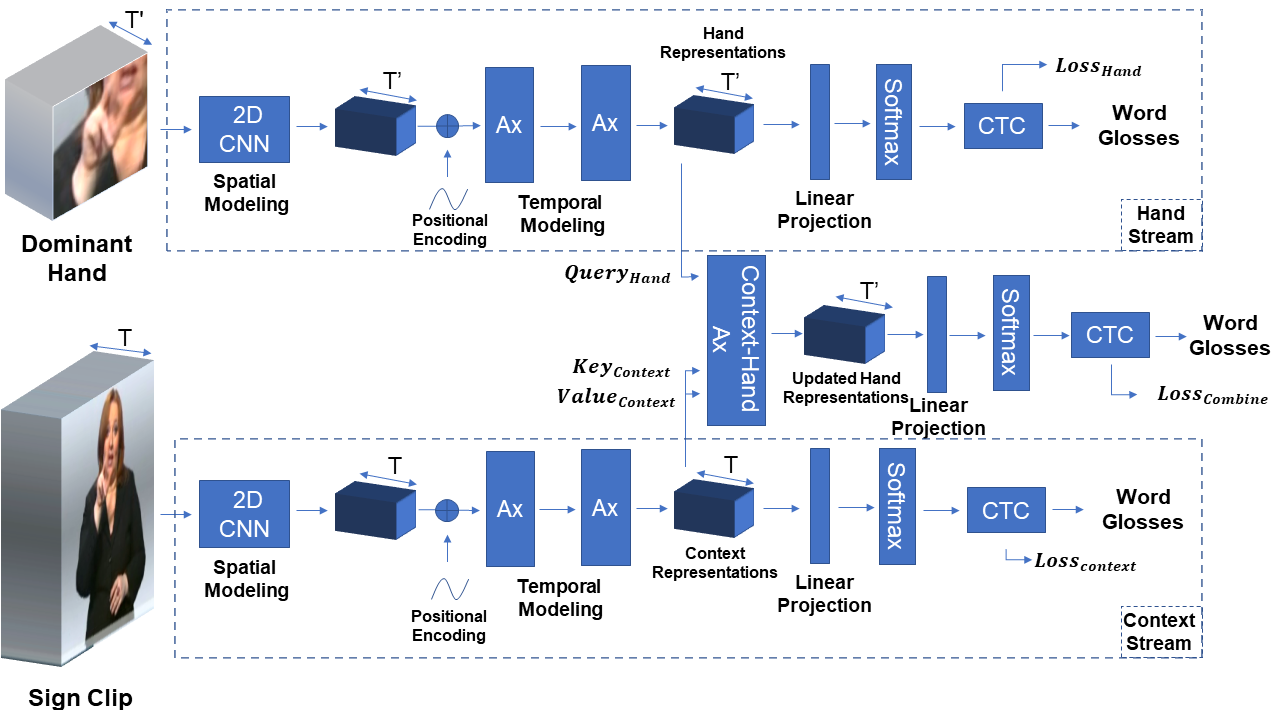}
 \caption{Combination of both the full-frame and the handshape streams through a Context-Hand Attention layer.}
 \label{fig:2}
\end{figure*}

Although our model may be able to learn all sign language modalities simply from the full-frame sequences, it would be of special interest to investigate the impact of fusing hand and global features. Handshapes require spatio-temporal information; as a result, each hand feature needs to attend to context across time and not just the current context frame.
Respectively, we design two-stream sub-networks: a Context Stream that is trained on the full-frame sequences and a Hand Stream which is trained on the cropped hand images. This will allow the first to learn to recognize global information and as a result the overall context of the signs. The second will be equipped to merely learn the handshape information. We combine the two through a third module that we refer to as a Context-Hand Attention Layer, in which the key and value features of shape $(T \mathsf{x} d_k)$ are computed as linear projections of the full-frame sequence representation, while the Query features $Q_{hand}$ of shape $(T'\mathsf{x} d_k)$ are obtained through the hand sequence projections. We apply the dot-product to get the attention values of the $Q_{hand}$ features over the $K_{context}$ features and then a weighted averaging of the resulting matrix over the $V_{context}$ features to get the updated hand representations. We apply a Layer-Norm operation on the hand query and then add it to the original hand features. The resulting feature is passed to another Norm Layer and a 2 layer Feed-Forward Network, to eventually produce the final hand query. The updated hand features are passed to a linear layer followed by a Softmax and then to a CTC to generate the word glosses. Similarly to \cite{camgoz2017subunets}, the overall network is trained end-to-end using the three loss layers: $\mathcal{L}_{hand}$, $\mathcal{L}_{context}$ and $\mathcal{L}_{combine}$. The described process and the overall architecture is illustrated in Figure \ref{fig:2}.

\subsection{Relative Local Context Masking}
	
Typically, self-attention allows the model to observe the entire sequence when measuring the attention at each sequence step. This can be very helpful for recognition, as Sign Language possesses a complex temporal structure and having this type of attention can help capture entangled dependencies. However, while handshapes expect temporal context, it is unclear to what extent this context may span. Required information can be limited to a local temporal window. So, instead of taking information from the entire sequence, we set a relative window and we only attend to the context around the current handshape frame. A similar idea has been explored for Acoustic Models \cite{sperber2018self}. This can be especially beneficial when dealing with long sequences where early handshapes don't usually require information from distant context frames.  We set a relative window of size $r$ and we apply a mask $M_{rel}\in \!R^{T'\mathsf{x}T}$ to the attention scores to mask out unwanted values in which:
    
$$
M_{rel} = \begin{cases}
    1 & \text{if $ |j-k| < r$}\\
    0 & \text{else}\\
\end{cases}
$$
	
Having attention restricted to a local region may result in missing some required long-range dependencies. Therefore, we avoid using such a mechanism on self-attention layers in both the context and hand streams as they are responsible for learning global information from the overall sequences. And we solely use it in the Context-Hand attention layer to incorporate handshapes to their local related contexts. 

\subsection{Implementation Details}
We start by extracting T frames (mostly 64) from the original video clip, in order to reduce computational complexity and discard frame redundancy. We resize the input full-frame images to a spatial resolution of 224 x 224. We also resize the hand patches to a lower resolution of 112 x 112. We normalize both full-frame and hand images by subtracting the dataset's image mean. For spatial image embeddings, we utilize the MobileNetV2 \cite{sandler2018mobilenetv2} architecture, due to its low-latency and high efficiency as a feature extractor. We drop its last fully connected layers and use the rest of the layers for feature embeddings. All our networks use a feature dimensionality of $d_k = 128$, 10 attention heads, and 2 layers of self-attention. The rest follow the default recommended setup provided by \cite{vaswani2017attention}: A 2 layer position-wise feed-forward network with a dimensionality of $d_{ff} = 2,048$ and a positional encoding sinusoidal function that encodes relative and absolute information. We based our networks' implementations on \cite{opennmt} and they are made publicly available\footnote{https://github.com/faresbs/slrt}.

\subsection{Training Details}

We initialize all of our networks' layers using Xavier \cite{glorot2010understanding}, except for the image embedding layers which have been pre-trained on ImageNet \cite{deng2009imagenet}. We use the Adam \cite{kingma2014adam} optimization method with its default parameters and a learning rate of $10^{-4}$.  We also use gradient clipping with a threshold of 1. We avoid overfitting by employing data augmentation through random x-y translation and using a dropout probability of $0.3$. We train all our networks using a small batch size of 2. Since we are performing batch training and considering that input clips may vary in size, sequences in every batch are padded to equal lengths (maximum sequence length in the batch). As a result, we utilize a mask on the input sequences to avoid attending on padded elements. In the case where we use relative local masking, we merge both the padding mask with $M_{rel}$. So that, besides the distant elements, we also avoid looking at the padded elements. 
We train our networks for 100 epochs or until train perplexity convergence. We evaluate our model every epoch on the validation set and report the best performing model.

\section{Experiments}

In this section, we evaluate our models on the task of Continuous Sign Language Recognition and we compare our findings with state-of-the-art previous works on the RWTH-PHOENIX-Weather 2014 corpus. The dataset offers sign clips (sequence of frames) and their corresponding gloss annotation. It contains 9 different signers and a train set of 5,672 sequences, thereby helping with model generalization. The textual annotation consists of a vocabulary $L$ of 1,231, with only 410 singletons (words that appear only once in the training set). The dataset provides full-frames alongside the cropped dominant hand (right hand), which provide a favourable test bed for our approach. Accordingly, we empirically demonstrate the effectiveness of combining the handshape features with their proper context using our proposed method, and we show that it significantly improves recognition performance. 

\subsection{Quantitative Results}

As shown in Table \ref{table:t2}, we start by comparing our approaches' results for the RWTH-PHOENIX-Weather 2014 dataset on both the validation and test sets and using the Word Error Rate: 

\begin{equation}
WER = \frac{\#\text{deletions} + \#\text{insertions} + \#\text{substitutions}}{\#\text{number of reference observations}}
\label{eq:5}
\end{equation}

Instead of using a greedy approach for decoding, in which we simply take the word with the highest probability, we consider using a CTC beam decoder with a beam width of 10 to get our final output sequence in both training and evaluation. We found that having a higher beam value did not improve performance and rather tremendously increased decoding time. Taking this into account, we use the same beam search strategy for all our experiments.  

\begin{figure}[t]
    \centering
    \includegraphics[width=0.4\textwidth]{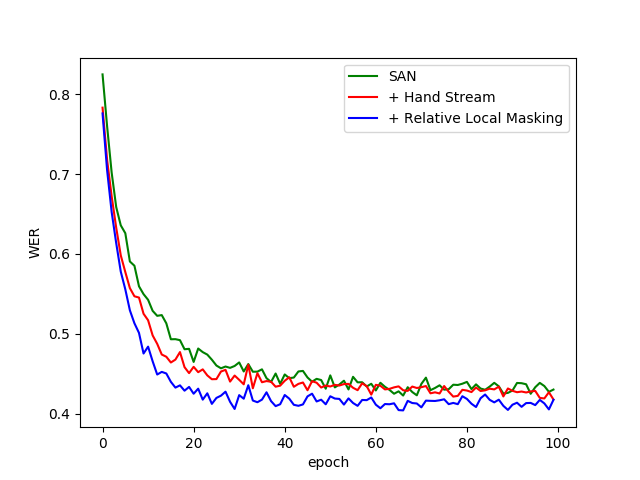}

    \caption{The Word Error Rate learning curve of our SAN variants for the task of CSLR on the RWTH-PHOENIX-Weather 2014 dataset.}
    \label{fig:learning_curve}
\end{figure}

First, we study the effect of incorporating handshape features with full-frame context information by adding a hand stream module and a Context-Hand attention layer. 
This leads to a considerable boost in performance as shown in Table \ref{table:t2}. But more importantly and as manifested in Figure \ref{fig:learning_curve}, adding handshape features also improves training and accelerates model convergence. This empirically showcases the usefulness of combining the dominant hand with the overall context derived from the nonmanual components of the sign.
Note that both hand and full-frame sequences have equal lengths $(T = T')$. Next and as demonstrated in Table \ref{table:t2}, limiting attention and fusing handshapes with their local related context leads to a significant improvement in performance, surpassing all our previous approaches. This supports the idea that the handshape, when accompanied by its proper context, can notably help in recognition.






In view of the significant complexity of our SAN architecture and the training dataset being too limited, our proposed model may not be able to generalize well for the task in hand and can as a result perform badly in the evaluation set. One way to tackle this problem is to provide a better initialization scheme for our model by firstly training the spatial feature extractor (CNN) on the same dataset. The CNN module will be at first initialized on ImageNet and then trained on the RWTH-PHOENIX-Weather 2014 dataset. As can be seen in Table \ref{table:t_pretraining}, by pre-training the CNN module of our SAN model on the same dataset, we significantly outperform the equivalent model when merely pre-trained on ImageNet.

\begin{table}[t!]
\begin{center}

\setlength{\tabcolsep}{10pt}

\caption{Comparison of our Sign Attention Network variants on RWTH-PHOENIX-Weather 2014 in Word Error Rate \% (the lower the better).}
\label{table:t2}
\begin{tabular}{l | c c}
& Dev & Test
\\

\hline

SAN & 35.33 & 35.45 \\

+ Hand Stream & 33.68 & 34.12 \\

\textbf{+ Relative Local Masking} & \textbf{32.74} & \textbf{33.29} \\

\hline
\end{tabular}
\end{center}
\vskip -0.2in 
\end{table}


\begin{table}[t!]
\begin{center}

\setlength{\tabcolsep}{10pt}

\caption{SAN recognition performance in Word Error Rate when pre-training its CNN module on ImageNet and RWTH-PHOENIX-Weather 2014 dataset \% (the lower the better).}
\label{table:t_pretraining}
\begin{tabular}{c | c c}
Pre-Training & Dev & Test
\\

\hline

ImageNet & 32.74 & 33.29 \\

\textbf{RWTH-PHOENIX-Weather 2014} & \textbf{29.02} & \textbf{29.78} \\

\hline
\end{tabular}
\end{center}
\vskip -0.2in 
\end{table}


\begin{table}[t]
\begin{center}

\setlength{\tabcolsep}{9pt}

\caption{Comparative results on the RWTH-PHOENIX-Weather 2014 dataset in Word Error Rate \% (the lower the better). 
}
\label{table:t3}
\begin{tabular}{l | c | c c}

& Dev & Test
\\


\hline

\textbf{SAN} & \textbf{29} & \textbf{29.7}  \\

Koller et al. (CNN-2BLSTM) \cite{koller2017re} & 32.7 & 32.9  \\

Koller et al. (CNN) \cite{koller2017re} & 33.7 & 33.3  \\

Huang et al. \cite{huang2018video} & 
- & 38.3 \\

Cui et al. \cite{Cui_2017_CVPR} & 39.4 & 38.7 \\

Koller et al. \cite{koller2016deep}  & 	38.3 & 38.8 \\

Camgoz et al. (HMM-LM) \cite{camgoz2017subunets} & 40.8 & 40.7 \\

Camgoz et al. (CTC) \cite{camgoz2017subunets} & 43.1 & 42.1 \\

Koller et al. \cite{koller2016deephand} & 47.1 & 45.1 \\

Koller et al. \cite{koller2015continuous} & 57.3 & 55.6 \\

\hline
\end{tabular}
\end{center}
\end{table}


Table 3 shows a comparison of our best performing model with previous publications. For a fair comparison,  we only compare with competing methods that are trained on ground truth target glosses and we don’t account for approaches that rely on forced alignment to train on per-frame labels. Accordingly, we opted to compare in this study with methods that employ similar training configurations, in order to showcase the efficiency of our proposed approach.

\begin{figure*}[t!]
    \centering
    \subfigure{\includegraphics[width=0.118\textwidth]{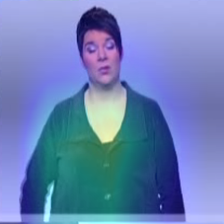}} 
    \subfigure{\includegraphics[width=0.118\textwidth]{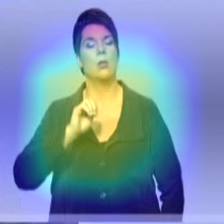}}
    \subfigure{\includegraphics[width=0.118\textwidth]{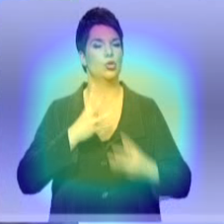}}
    \subfigure{\includegraphics[width=0.118\textwidth]{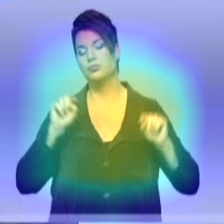}}
    \subfigure{\includegraphics[width=0.118\textwidth]{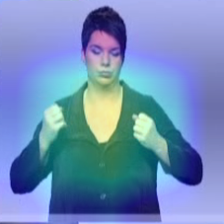}}
    \subfigure{\includegraphics[width=0.118\textwidth]{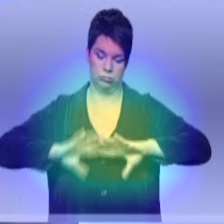}}
    \subfigure{\includegraphics[width=0.118\textwidth]{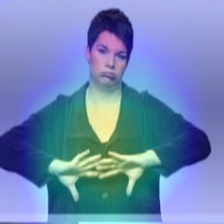}}
    
    \subfigure{\includegraphics[width=0.118\textwidth]{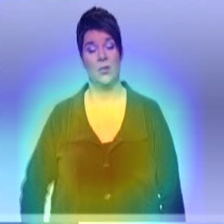}} 
    \subfigure{\includegraphics[width=0.118\textwidth]{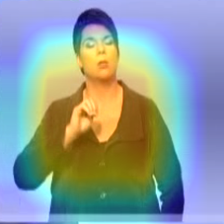}}
    \subfigure{\includegraphics[width=0.118\textwidth]{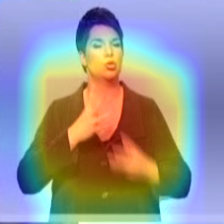}}
    \subfigure{\includegraphics[width=0.118\textwidth]{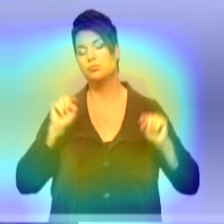}}
    \subfigure{\includegraphics[width=0.118\textwidth]{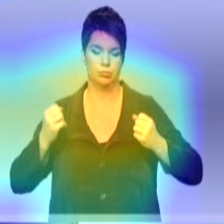}}
    \subfigure{\includegraphics[width=0.118\textwidth]{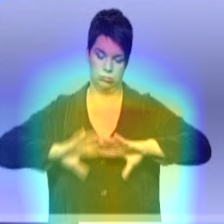}}
    \subfigure{\includegraphics[width=0.118\textwidth]{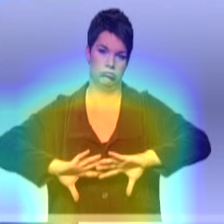}}
    
    \subfigure{\includegraphics[width=0.118\textwidth]{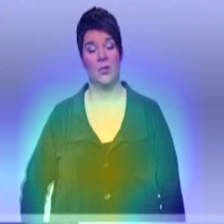}} 
    \subfigure{\includegraphics[width=0.118\textwidth]{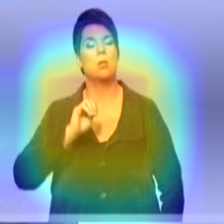}}
    \subfigure{\includegraphics[width=0.118\textwidth]{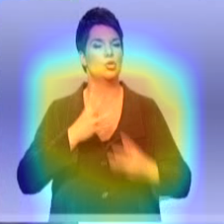}}
    \subfigure{\includegraphics[width=0.118\textwidth]{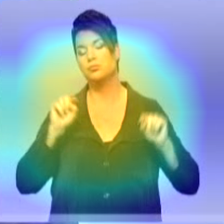}}
    \subfigure{\includegraphics[width=0.118\textwidth]{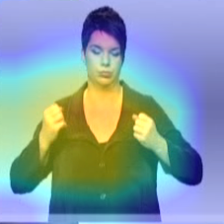}}
    \subfigure{\includegraphics[width=0.118\textwidth]{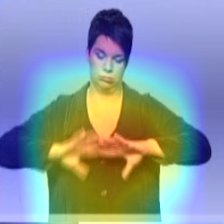}}
    \subfigure{\includegraphics[width=0.118\textwidth]{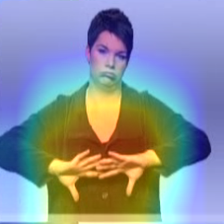}}

    \caption{Heatmap Localization on the frame embedding activations, highlighting important regions which the model uses to predict a particular sign. The top sequence is the output results of our SAN network. The middle is for SAN with the hand stream and the bottom is for SAN with hand stream and the local context masking. Note that this example is randomly chosen and not cherry-picked.}
    \label{fig:heatmap}
\end{figure*}

\subsection{Qualitative Analysis}

Apart from obtaining promising results, it is of interest to visualize the areas on which the model focuses for sign prediction. Accordingly, we use Grad-Cam \cite{selvaraju2017grad} to produce a localization heatmap in which bright pixels represent positive influence and have great importance on the predicted sign. As can be seen in Figure \ref{fig:heatmap}, our models primarily focus on the dominant hand (right hand) and the face area which reinforces the intuition that our model is able to identify the essential components for sign interpretations.

\section{Conclusion and Future Work}

In this work,  we have proposed a novel method that exploits attention to effectively combines hand query features with their respective temporal full-body context without the need for any additional supervision. We have proven the efficiency of such an approach to the task of Continuous Sign Language Recognition. For future studies, it will be of interest to investigate the effect of using a forced alignment algorithm on top of our architecture, similarly to \cite{koller2017re, pami2019}. Relying on forced alignment may significantly improve recognition as shown in \cite{koller2017re} and it is a popular solution to overcome weak supervision by iteratively refining and training on label-to-image prediction. We can also employ HMMs instead of CTC for sequence alignment, as they have been proven to be superior in \cite{camgoz2017subunets}. Another important venue to explore is to furthermore extend this work by applying our architecture for the task of Sign Language Translation (SLT) similarly to \cite{cihan2018neural} and \cite{camgoz2020} and investigating the effect of combining hand features with their global non-manual context through the use of the attention mechanism.

\ifCLASSOPTIONcompsoc
\section*{Acknowledgments}
\else
  \section*{Acknowledgment}
\fi

The authors gratefully thank the reviewers and the associate editor for their valuable comments and important suggestions. This work is supported by research grants from the Natural Sciences and Engineering Research Council of Canada (NSERC).






%



\balance 
\bibliographystyle{IEEEtran.bst}
\bibliography{bare_conf}

\end{document}